\documentclass[pdflatex,sn-mathphys-num]{sn-jnl}% Math and Physical Sciences Numbered Reference Style

\usepackage{graphicx}%
\usepackage{multirow}%
\usepackage{amsmath,amssymb,amsfonts}%
\usepackage{soul}
\usepackage{amsthm}%
\usepackage{mathrsfs}%
\usepackage[title]{appendix}%
\usepackage{xcolor}%
\usepackage{textcomp}%
\usepackage{manyfoot}%
\usepackage{booktabs}%
\usepackage{algorithm}%
\usepackage{algorithmicx}%
\usepackage{algpseudocode}%
\usepackage{listings}%

\theoremstyle{thmstyleone}%
\theoremstyle{thmstyletwo}%

\theoremstyle{thmstylethree}%

\begin{document}

\title[Article Title]{A Hybrid Model-based and Data-based Approach Developed for a Prosthetic
Hand Wrist}


\author*[1]{\fnm{Shifa} \sur{Sulaiman}}\email{ssajmech@gmail.com}

\author[1]{\fnm{Francesco} \sur{Schetter}}\email{schfrn@gmail.com}

\author[2]{\fnm{Mehul} \sur{Menon}}\email{menonm98@gmail.com}

\author[1]{\fnm{and Fanny} \sur{Ficuciello}}\email{fanny.ficuciello@unina.it}

\affil[1]{\orgdiv{Department of Information Technology and Electrical Engineering}, \orgname{University of Naples Federico II}, \orgaddress{\street{Claudio}, \city{Naples}, \postcode{80125}, \state{Campania}, \country{Italy}}}

\affil[2]{\orgdiv{Department of Mechanical Engineering}, \orgname{National Institute of Technology}, \orgaddress{\street{Mahatma Gandhi Avenue}, \city{Durgapur}, \postcode{713209}, \state{West Bengal}, \country{India}}}


\abstract{ The incorporation of advanced control algorithms
into prosthetic hands significantly enhances their ability to
replicate the intricate motions of a human hand. This work
introduces a model-based controller that combines an Artficial Neural Network (ANN) approach with a Sliding Mode
Controller (SMC) designed for a tendon-driven soft continuum
wrist integrated to a prosthetic hand known as ’PRISMA
HAND II’. Our research focuses on developing a controller
that provides a fast dynamic response with less computational
effort during the motions of the wrist. The proposed controller
consisted of an ANN for computing bending angles incorporated
along with an SMC to regulate tendon forces. Kinematic and
dynamic models of the wrist are formulated using Piece-wise
Constant Curvature (PCC) hypothesis. The performance of
the proposed controller is compared with other controller
strategies developed for the same wrist. Simulation
studies and experimental validations of motions of the fabricated wrist employing the controller are included in the paper. 

}

\keywords{Prosthetic hand, Soft robot, Hybrid controller, Data based approach, Model based controller}

\maketitle

\section{Introduction}\label{sec1}
The application of control strategies to regulate motions of prosthetic hands is essential for improving functionality and usability of these advanced devices \cite{fanny1}, \cite{fanny2}. By integrating sophisticated control algorithms, prosthetic hands can more accurately mimic the natural movements of a human hand, allowing users to perform a wider range of tasks with greater precision. 
These advanced control strategies often utilize sEMG or force sensors to detect muscle activity or interaction forces, which are interpreted by reinforcement learning-based controllers to adaptively generate grasping motions or trajectory adjustments in real time \cite{de}.
Effective control strategies are particularly important to individuals who rely on prosthetic limbs for daily activities, as effective control strategies can significantly improve their quality of life \cite{control1,control2} . 
The development of adaptive and intuitive control systems enables users to engage in complex tasks, such as grasping objects of varying shapes and sizes, thereby enhancing the overall functionality of the prosthetic device \cite{palli},\cite{palli1}. The synergy between kinematic and dynamic models in controller design not only improves the precision of soft robots but also expands their functional capabilities \cite{review}. 
The integration of these modeling strategies enables the development of controllers that can effectively manage the inherent uncertainties associated with soft materials. 
Hybrid controller strategies incorporating  Artificial Neural Network (ANN) based algorithms are developed nowadays to improve the performance of the soft robotic motions with less computational efforts \cite{ANN}.

This study presents a control scheme that integrates an Artificial Neural Network (ANN) with a Sliding Mode Controller (SMC) to manipulate motions of a soft continuum wrist attached to a prosthetic hand. We combined the adaptive learning capability of ANN for modeling complex, nonlinear system dynamics with the robustness of sliding mode control (SMC) to handle external disturbances and model uncertainties. We adopted Piece-wise Constant Curvature (PCC) modelling strategy to determine kinematic and dynamic models of the soft continuum wrist. PCC offered a solution to the complexities associated with the infinite dimensionality due to the complex configurations of soft robots. The model obtained using PCC was used in the controller strategy to increase accuracy and performance of the controller outputs. 
 The major contributions of the presented work are as follows:
\begin{itemize}
    \item Modelling of a soft continuum wrist based on a 4-element parameterized configuration using PCC method.
    \item Development of a data and model-based approach combining ANN and SMC for improving the performance of wrist motions.
    \item A comparison study of the proposed controller with a PID controller and other model-based controllers to demonstrate the advantages of the proposed controller.
    \item An experimental validation proving the effectiveness of the proposed controller in presence of external disturbances during real-time implementations with reduced computational effort. 
\end{itemize}
The outline of this paper is given as follows: Section II presents related work on the development of controllers for soft continuum mechanisms. Design and modelling procedures are given in section III. SMC strategy is demonstrated in section IV. Section V presents the simulation and experimental results. Section VI presents the conclusion of this work, summarizing key findings and outlining potential directions for future research.

\section{BACKGROUND}
The integration of tendons allows soft robots to achieve complex motions while maintaining structural integrity. 
%This approach enhances the overall functionality and responsiveness of soft robotics. 
Tendon-driven soft continuum robots offer several notable advantages that enhance functionality and versatility of the whole mechanism 
%Their design allows for a high degree of flexibility and adaptability, enabling them to navigate complex environments and perform intricate tasks. 
%Design of soft continuum robots actuated using tendons were presented in %\cite{orekhov2017modeling}, %\cite{wooten2018vine}, 
\cite{wockenfuss2022design}, \cite{tutcu2021quasi}. Various modeling techniques such as Constant Curvature (CC) models \cite{escande},  Variable Curvature (VC) models \cite{huang}, Finite Element Method (FEM) \cite{xavier}, Machine Learning (ML) methods \cite{renda} are employed to determine both static and dynamic models of soft robots. 

The major benefit of PCC modeling is its ability to accommodate the inherent flexibility and adaptability of soft robots \cite{ANN1}. %This approach allows for a simplified representation of the robot's shape and movement, facilitating easier calculations and simulations. 
Unlike traditional rigid-body models, which often struggle to accurately depict the dexterous motions of soft robots, the PCC model captures the continuous nature of soft robots while maintaining computational efficiency.

PCC-based modeling has been widely adopted for soft continuum robots due to its ability to approximate complex, continuous deformations using discrete constant-curvature segments. This approach enables simplified kinematic formulations while maintaining sufficient accuracy for many practical applications. By reducing the complexity of continuous deformation into manageable geometric primitives, PCC facilitates real-time control and simulation, which is particularly valuable for robots operating in constrained or dynamic environments. Moreover, it allows for modular representation of multi-segment structures, making it easier to integrate with optimization and learning-based control frameworks. PCC models are also compatible with various actuation strategies including tendon-driven, pneumatic, and hybrid systems further enhancing their versatility across different soft robotic platforms. Despite its approximative nature, PCC has proven effective in capturing essential motion characteristics, especially when combined with empirical calibration or adaptive control techniques.

PCC-based modelling of a rigid-soft continuum robot actuated using bellows was carried out in \cite{r1}. Experimental validation showcased a higher positional accuracy (higher than 98\%) during the motions of the robot. A PCC method employed for deriving kinematic model of a soft robot manufactured using latex rubber was demonstrated in \cite{r2}.  Experimental validations demonstrated that the model could not sustain load when the loading surpassed the stiffness of the trunk. PCC based 3D modelling of a tendon-driven soft continuum robot considering twisting effects was presented in \cite{rao}. The static modelling methodology was employed for modelling a tendon-driven continuum robot that encounters stability issues due to dimensional constraints on the rods along their backbone. 
A continuum robot modelled using an approximate piecewise constant curvature (APCC) method was illustrated in \cite{cheng}. The APCC model simplified the original kinematic model by employing classical rigid linkages, thereby enhancing numerical stability. 
Another soft continuum robot modelled using PCC method incorporating a 4 parameter configuration criteria was discussed in \cite{della}. A dynamic model based controller was implemented employing the PCC method.

Incorporating kinematic and dynamic models into a controller scheme facilitates a structured and efficient way to handle curvature-related computations \cite{ANN2}. SMC strategies are presently employed in a wide variety of applications to control the movements of various soft robots owing to its superior performance capabilities \cite{ANN3}.  
An adaptive SMC developed for a soft robotic actuator was presented in \cite{cao}. The proposed control strategy showcased tracking precision utilizing a non-singular sliding surface and a higher-order sliding mode observer. An adaptive SMC strategy, that demonstrated resilience to uncertainties in model parameters and unanticipated input disturbances was illustrated in \cite{Kazemipour}. Experiments conducted using a soft continuum arm to assess the controller's effectiveness in tracking task-space trajectories across various payloads exhibited a tracking accuracy more than 98\% compared to a traditional controller. 
The performance of a feed forward SMC scheme for controlling a soft pneumatic actuator was evaluated in \cite{Skorina}. The controller scheme aided the robot in following dynamic trajectories across various frequencies in the presence of an external force with reduced error values and chattering.
In \cite{khan}, a soft robot was controlled using an SMC with a PID sliding surface. The suggested controller employed feedback error to establish a PID sliding surface, while a nonlinear SMC ensured that the system aligned with this surface. 
%The parameters of the PID sliding surface influenced the dynamic characteristics of the soft robot. 
The experimental findings indicated that the suggested controller effectively reduced vibration amplitude to a significantly lower range. SMC scheme of a tendon- driven soft robot employing cosserat rod theory was presented in \cite{Mousa}. A generalized $\alpha$ method was adopted for reducing computational load during analyses.

The integration of artificial neural networks ANN with SMC enhances system adaptability to nonlinear dynamics while maintaining robustness against external disturbances and model uncertainties. Firstly, the integration of ANN allows for the approximation of complex nonlinear functions, which can effectively model system dynamics that are difficult to characterize using traditional methods \cite{Al}. This capability enables the controller to adapt to changing system conditions and uncertainties, thereby improving robustness and stability.
Furthermore, the combination of these two methodologies can lead to improved convergence rates and reduced chattering effects, which are common challenges in conventional SMCs.

A comprehensive review of recent literature on soft continuum robot control conducted as part of this work identified several recurring limitations in existing controller strategies, including sub-optimal control performance, high computational demands, inaccurate modeling assumptions, and challenges in tuning parameters. This paper outlines the techniques and approaches utilized to develop a model and data-based scheme for a controller, that is capable of controlling the motions of a soft wrist in presence of external disturbances with improved precision and reduced computational effort.

\section{DESIGN AND MODELLING OF THE SOFT CONTINUUM WRIST}
This section presents the components and mechanical design of the wrist integrated into the prosthetic hand. Kinematic and dynamic modellings of the wrist using PCC method are also included in this section.

\subsection{Design of the Wrist}
\begin{figure}[hbt!]
\centerline{\includegraphics[width=0.60\textwidth, height = 1.8 in]{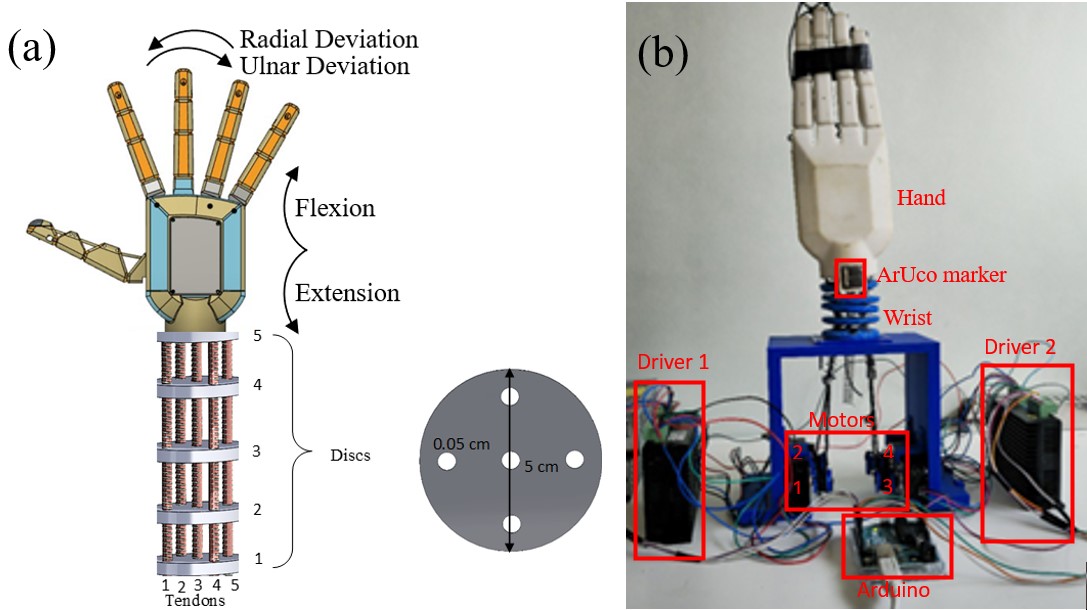}}
\caption{Wrist section (a) Conceptual model (b) Fabricated model}
\label{conceptual model}
\end{figure}

We developed a soft continuum wrist section \cite{conf} and integrated it with the prosthetic hand known as PRISMA HAND II \cite{ref2}. The wrist was designed to meet specific load-bearing requirements and to support motion in four anatomical directions: ulnar deviation, radial deviation, flexion, and extension. The wrist integrated to the PRISMA HAND II along with the directions of motions are shown in Fig. \ref{conceptual model}(a). 
The wrist section was composed of 5 rigid discs, 5 springs, and 5 tendons inserted through the springs. The tendons are manipulated to obtain desired motions in different directions \cite{conf}. Due to the arrangement of the components, the wrist section can also compress and expand in axial direction. 
With regard to disc 5, the wrist's range of motion was set to -50$^{\circ}$ to +50$^{\circ}$ in all directions. This was adequate for the PRISMA HAND II to perform manipulation tasks.

\subsection{Kinematic modelling of the wrist using PCC method}
\begin{figure}[hbt!]
\centerline{\includegraphics[width=1.0\textwidth, height = 1.6 in]{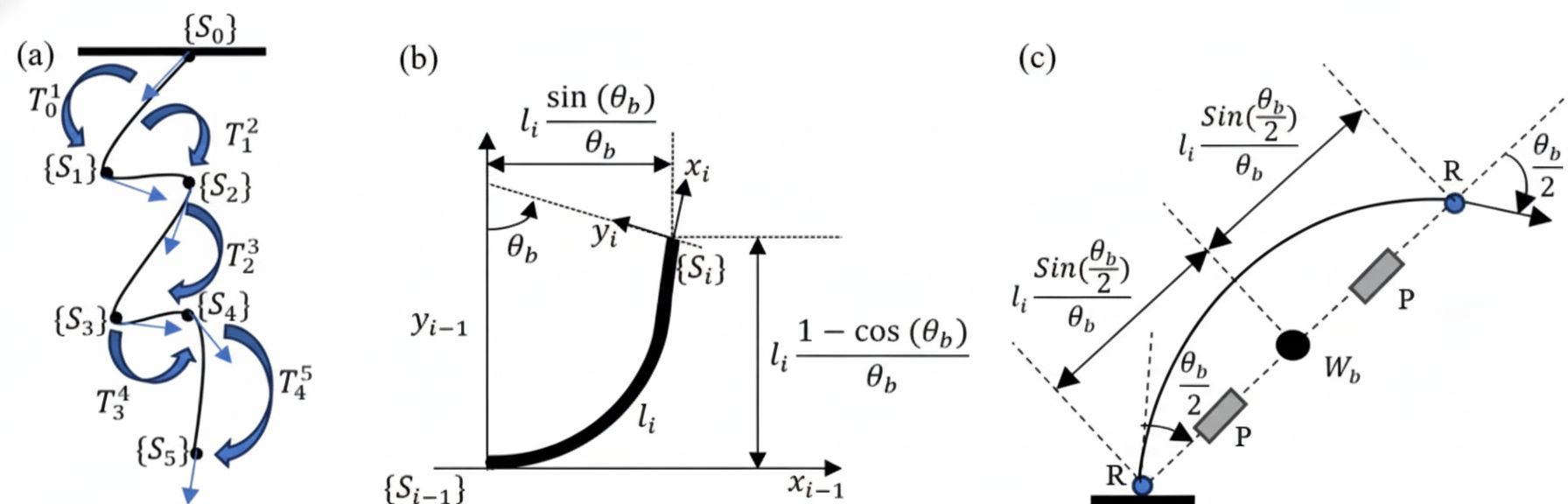}}
\caption{Configuration (a) Transformations (b) Single segment (c) RPPR}
\label{conf}
\end{figure}

In the context of soft robotics, the PCC model provides a systematic framework for understanding the robot's configuration by breaking it down into discrete segments that exhibit CC. The segmentation is crucial, as it simplifies the complex nature of soft robots, which often possess an infinite number of degrees of freedom due to their flexible and deformable structures. By employing a finite number of segments, the model not only enhances computational efficiency but also allows for a more intuitive analysis of the robot's kinematics and dynamics. Consider a soft continuum robot with $n$ CCs shown in Fig. \ref{conf}(a) and corresponding frames attached to the end of each segment are represented as $S_{0},S_{1},.....,S_{n}$. Transformation matrices from $n^{th}$ frame to zeroth frame are represented as  $^{0}T_1,^{1}T_2,......,^{n-1}T_n$. 
A single section of the soft continuum robot is shown in Fig. \ref{conf}(b). The robot currently in $i^{th}$ plane consisted of two inertial frames at the ends, namely $S_{i-1}$ and $S_{i}$. Bending angles, $\theta_b$ were obtained by considering a relative rotation between the two inertial frames of the segment. The 2D homogeneous transformation matrix of the soft robot, $^{i-1}T_i $ with section length, $l_i$ from $i-1^{th}$ frame to $i^{th}$ frame in $i^{th}$ plane is given in equation \ref{eq:1}.
\begin{equation}
    ^{i-1}T_i = 
    \begin{bmatrix} 
        cos\theta_b & -sin\theta_b &l_i\frac{sin\theta_b}{\theta_b} \\
        sin\theta_b &cos\theta_b &l_i\frac{1-cos\theta_b}{\theta_b} \\
        0 &0 &1
    \end{bmatrix}
    \label{eq:1}
\end{equation}
In this work, we present a framework that establishes an equivalence between the modelling of a soft robot and a rigid robot using DH parametrization \cite{della}. 
%In \cite{r3}, an alternative representation of equation \ref{eq:1} was presented, utilizing elemental Denavit-Hartenberg (DH) transformations. 
We can generate conceptual linkages correspond to a PCC soft continuum robot using rigid robot topologies. The state-space representation of an equivalent rigid robot can be used for creating an augmented representation of a PCC soft continuum robot. Consider $\mu \in R^{n~ m}$ as the augmented configuration of a PCC soft robot with $m$ number of joints per CC. The two configurations of the robot are mapped as follows:
 \begin{equation}
     \mu:R^{n}\rightarrow R^{n~m}
 \end{equation}
The mapping ensures that the terminal points of each CC segment align precisely with the corresponding reference points of the rigid robotic structure. From a kinematic perspective, any augmented representation that fulfills the aligning condition can be deemed equivalent. This equivalence is crucial for maintaining the integrity of the robotic motion and ensuring that the kinematic relationships are preserved across different configurations. 
Consider an equivalent rigid robot shown in Fig. \ref{conf}(c) with RPPR (R:Revolute, P:Prismatic) configuration. A point mass, $W_b$ positioned at the center of the main chord serves as an appropriate approximation for the mass distribution of the CC segment. DH parametrization of the equivalent rigid body is given in table \ref{table:DH}. $\theta_{dh}$, $d$, $a$, and $\alpha$ are classic DH parametrization. 
\begin{table}[hbt!]
 \caption{DH parametrization table of equivalent rigid robot}
    \centering
    \begin{tabular}{|c|c|c|c|c|c|}    
    \hline
     Links &$\theta_{dh}$ &$d$ &$a$ &$\alpha$ &Mass \\
         \hline
        1 & $\theta_{b}/2$ &0 &0 & $\pi/2$ &0\\
        2 & 0 & $l_i \frac{\sin(\theta_b/2)}{\theta_b}$ & 0 & 0 & $W_b$ \\ 3 & 0 & $l_i \frac{\sin(\theta_b/2)}{\theta_b}$ & 0 & $-\pi/2$ & 0 \\ 4 & $\theta_b/2$ & 0 & 0 & 0 & 0 \\ \hline
    \end{tabular}
    %\caption{DH parametrization table of equivalent rigid robot}
    \label{table:DH}
\end{table}
The configuration of the PCC modelled soft wrist obtained using mapping between the augmented rigid configuration and CC segment is given in equation \ref{eq:2}.
\begin{equation}
\mu(\theta_{b}) = [\mu_1(\theta_{b1})^{T} \mu_2(\theta_{b2})^{T} .............. \mu_n(\theta_{bn})^{T}]^T
    \label{eq:2}
\end{equation}
where $\mu_i(\theta_{bi})$ is the 4-element parameterized configuration of an individual segment obtained using DH parametrization and given in equation \ref{eq:3}.
\begin{equation}
\mu_i(\theta_{bi}) =
\begin{bmatrix}
    \frac{\theta_{bi}}{2}
    &l_i\frac{sin\frac{\theta_{bi}}{2}}{\theta_{bi}}
   & l_i\frac{sin\frac{\theta_{bi}}{2}}{\theta_{bi}}
    &\frac{\theta_{bi}}{2}
\end{bmatrix}
^T
    \label{eq:3}
\end{equation}
Transpose of Jacobian of these matrices with respect to $\theta_b$, $J_\mu(\theta_b)$ = $\frac{\partial \mu}{\partial \theta_b}$ was used in dynamic modelling procedure for mapping the wrist section configuration to the 4 parameter configuration as explained in the section 3.3. 
\subsection{Dynamic modelling of the wrist}
The generalised dynamic model of the wrist with respect to generalised joint parameter, $\theta$ and their derivatives $(\dot \theta, \ddot \theta)$ is given in equation \ref{eq:4}.
\begin{equation}
M_b(\theta)(\ddot \theta) + (C_b+D_b)(\theta, \dot \theta)(\dot \theta) + G_b(\theta) + K_b(\theta) \theta = \tau_b + J^T_b(\theta) F_b
    \label{eq:4}
\end{equation}
where $M_b(\theta)$ is the $4n\times 4n$ generalised inertia matrix, $C_b(\theta, \dot \theta)$ is the $4n\times 4n$ generalised centrifugal and Coriolis matrix, $G_b(\theta)$ is the $4n\times 1$ gravitational field, $\tau_b$ is the $4n\times 1$ mapped actuation torque, $D_b$ is the damping, and $K_b$ is the stiffness of the soft wrist. Jacobian, $J_b$ maps external force, $F_b$ to the joint torque values.
The dynamic model of the wrist after incorporating mapping Jacobian, $J_\mu: R^{n} \rightarrow R^{4n \times n}$ is given in equation \ref{eq:6}.
\begin{equation}
M(\theta)(\ddot \theta) + (C+D)(\theta, \dot \theta)(\dot \theta) + G(\theta) + K(\theta)\theta = \tau + J^T(\theta) F
    \label{eq:6}
\end{equation}
where $M(\theta)$ is the $n\times n$ mapped inertia matrix and given in equation \ref{eq:7}
\begin{equation}
    M(\theta) = J^T_\mu(\theta)~M_{b}(\mu(\theta))~J_\mu(\theta)
    \label{eq:7}
\end{equation}
$C(\theta, \dot \theta)$ is the $n\times n$ mapped centrifugal and Coriolis matrix given in equation \ref{eq:8}
\begin{equation}
C(\theta, \dot \theta) = J^T_\mu(\theta)M_{b}(\mu(\theta))\dot J_\mu(\theta,\dot \theta) + J^T_\mu(\theta)D_{b}(\mu(\theta),J_\mu(\theta)\dot \theta) J_\mu(\theta)
\label{eq:8}
\end{equation}
$G(\theta)$ is the $n\times 1$ mapped gravitational field given in equation \ref{eq:9}
\begin{equation}
    G(\theta) =  J^T_\mu(\theta)~G_{b}(\mu(\theta))
    \label{eq:9}
\end{equation}
$\tau$ is the $n\times 1$ mapped actuation torque given in equation \ref{eq:10}
\begin{equation}
    \tau = J^T_\mu(\theta)~\tau_{b}
    \label{eq:10}
\end{equation}
The relation between $\tau_b$, $n \times 1$ tendon force ($F_t$), $\tau_b$, and $n \times n_a$($n_a$:number of actuators) actuation matrix ($A_b$) is given in equation \ref{force}
\begin{equation}
\tau_b = A_b F_t
    \label{force}
\end{equation}
Tendon force, $F_t$ is obtained using the following equation \ref{f2}
\begin{equation}
F_t = [J^T_\mu(\theta)A_{b}]^{-1}\tau
    \label{f2}
\end{equation}
%where $\tau_b$ is the $n \times 1$ tendon force along with $n \times n_a$($n_a$:number of actuators) actuation matrix, $A_b$ given in equation \ref{f}
%\begin{equation}
    %\tau_b = J^T_\mu(\theta)~\tau_{b}
   %\label{eq:10}
%\end{equation}
Jacobian, $J$ is given in equation \ref{eq:11}
\begin{equation}
    J = J_{b}(\theta)~J_\mu(\theta)
    \label{eq:11}
\end{equation}
The dynamic model was employed in SMC strategy to increase the accuracy of control action. 

\section{SLIDING MODE CONTROLLER STRATEGY}
\begin{figure}[hbt!]
    \centering
\includegraphics[width=1\textwidth]{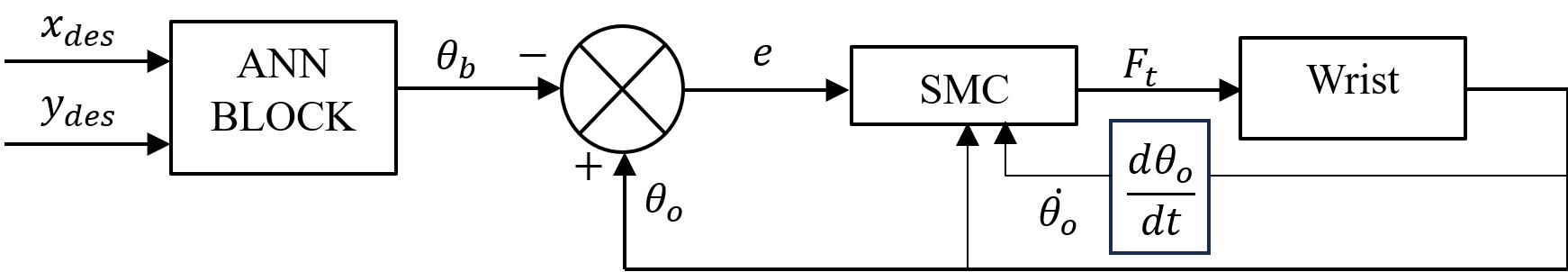}
   \caption{SMC strategy}
    \label{hybrid}
\end{figure}

\begin{figure}[hbt!]
\centerline{\includegraphics[width=0.4\textwidth, height = 1.8 in]{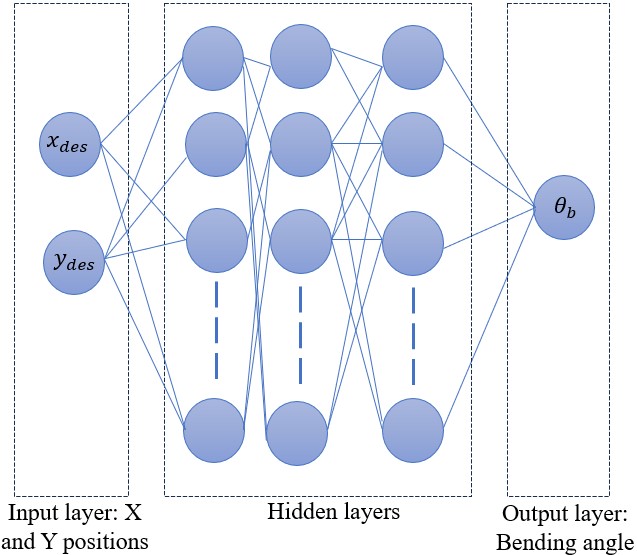}}
\caption{ANN architecture}
\label{ANN}
\end{figure}

A model based controller employing dynamic model was designed for controlling motions of the wrist section while traversing trajectories. The proposed controller can linearize and ensure a global asymptotic stability. 
A controller scheme (shown in Fig. \ref{hybrid}) was developed combining an ANN strategy and an SMC to manipulate the motions of the soft wrist integrated to the prosthetic hand. The proposed controller framework (shown in Fig. \ref{hybrid}) leverages the strengths of both ANN and SMC to achieve precise motion control of the soft wrist component attached to the prosthetic hand.
2D desired positions ($x_{des}, y_{des}$) were fed to a ANN block and corresponding bending angles, $\theta_b$ were obtained as the output. Actuation forces, $F_t$ were obtained as the output from SMC block and fed to wrist model. Output bending angles, $\theta_o$ were used for calculating errors, $e(t)$ as given in equation \ref{eq:12}
\begin{equation}
    e(t)=\theta_o -\theta_b
    \label{eq:12}
\end{equation}
$\theta_o$, $\dot \theta_o$, and $e(t)$ were given as the input to SMC block to correct the output motions. The implementation of the SMC served as an effective strategy for regulating the motions of the soft wrist due to its robustness against uncertainties and external disturbances \cite{ANN3}. %which are common in the operation of soft robots.
Consider the non-linear conceptual wrist system of the form including vector field function,$f(\theta)$ and force field function, $g(\theta)$ given in equation \ref{eq:13}
\begin{equation}
    \dot \theta = f(\theta) + g(\theta)U, \theta\in R^{n}, U\in R
    \label{eq:13}
\end{equation}
$f(\theta)$ is given in equation \ref{eq:14}
\begin{equation}
f(\theta) = 
\begin{bmatrix}
    \dot \theta_o \\
    M^{-1}[J^T F-(C+D)\dot \theta_{o} - K\theta_o - G]
\end{bmatrix}
\label{eq:14}
\end{equation}
$g(\theta)$ is the function which maps $U$ to the force field of the system as given in equation \ref{eq:15}
\begin{equation}
g(\theta) = 
\begin{bmatrix}
    0 \\
    M^{-1}
\end{bmatrix}
\label{eq:15}
\end{equation}
$U$ is the control input (in our case, $\tau$) combining equivalent component($U_{equivalent}$) and switching component($U_{switching}$) given in the equation \ref{eq:16}
\begin{equation}
    U = -[U_{equivalent} + U_{switching}]
    \label{eq:16}
\end{equation}
$U_{equivalent}$ is obtained based on the equation \ref{eq:17}
\begin{equation}
    U_{equivalent} = \frac {L_f(\sigma(\theta))}{L_g(\sigma(\theta))}
    \label{eq:17}
\end{equation}
where $L_f(\sigma(\theta)$ and $L_g(\sigma(\theta)$ are the lie derivatives of functions $f(\theta)$ and $g(\theta)$ with respect to $\sigma(\theta)$. $\sigma(\theta)$ is the sliding surface function given in equation \ref{eq:18}
\begin{equation}
    \sigma(\theta) = P_1e(t) + P_2\dot \theta_o
    \label{eq:18}
\end{equation}
where $P_1$ and $P_2$ are two tuning parameters to stabilise the controller and reduce the output errors to a tolerance range. The importance of tuning parameters in an SMC cannot be overstated, as these parameters play a crucial role in determining the performance and stability of the control system. Proper tuning of $P_1$ and $P_2$ ensures that the controller can effectively manage system dynamics, particularly in the presence of uncertainties and external disturbances. 
Sliding surface $S$ was selected based on the convergence in finite time to guarantee optimal control and is given in equation \ref{eq:19}
\begin{equation}
    S = {\dot \theta_o \in R^{n}: \sigma(\theta) = 0}
    \label{eq:19}
\end{equation}
$U_{switching}$ is given in equation \ref{eq:20}
\begin{equation}
    U_{switching} = \frac {1}{L_g(\sigma(\theta))} P_3~tanh(\sigma (\theta))
    \label{eq:20}
\end{equation}
where $P_3$ is a control parameter to increase the stability of the controller. $Tanh$ function was used instead of $sgn$ function and the generated trajectory was smoother compared to $sgn$ function. After determining $\tau$, $F_t$ was obtained using equation \ref{f2}. In order to maintain the system to the sliding surface, the following condition is applied as given in equation \ref{eq:21}
\begin{equation}
    u= 
    \begin{Bmatrix}
        u^+, &\sigma(\theta)>0 \\
        u^-, &\sigma(\theta)<0
    \end{Bmatrix}
    \label{eq:21}
\end{equation}
Along with SMC, implementation of an ANN algorithm to predict the desired bending angles aided in elevating the performance of the control scheme. The ANN network used for predicting bending angles based on the desired x and y positions of the last disc (disc 5) of the soft continuum wrist (Fig. \ref{ANN}).
From a total of 1,000 samples, 750 were utilized for training and 250 were reserved for testing and validating the outcomes. 3 hidden layers consisted of 200, 100, and 100 neurons respectively. Performance objective was set to zero and initial learning rate was assumed as 0.01. Along sigmoid function, we also trained the ANN using ReLU, Leaky ReLU, and Tanh functions.Hidden layer optimization was performed via grid search, testing architectures with 1 to 5 fully connected layers. Each configuration was evaluated using 5-fold cross-validation. The 3-layer architecture achieved the best trade-off between accuracy and computational efficiency, with diminishing returns observed beyond three layers. ADAM (Adaptive Moment Estimation) method \cite{ANN4}, an extension of the stochastic gradient descent (SGD) method with momentum was used as the backpropagation technique. Sigmoid function was chosen for neuron activation in the network following a comparative analysis with 3 other alternative activation functions. This decision was based on the performance metrics observed during the evaluation, which indicated that the sigmoid function provided superior results in terms of accuracy and stability.

\section{RESULTS AND DISCUSSION}
We have conducted both simulation and experimental validation of the proposed control algorithm on the wrist section of the system. The controller’s performance was assessed under two conditions: with and without the application of external forces. A comparative analysis between the proposed SMC scheme and other model-based controllers developed for the same wrist section is presented in the sections 5.1 - 5.3. Additionally, experimental trials were performed to evaluate the real-time effectiveness of the SMC in executing dynamic tasks.

\subsection{Simulation study}
\begin{figure}[hbt!]
    \centering
\includegraphics[width=1\textwidth]{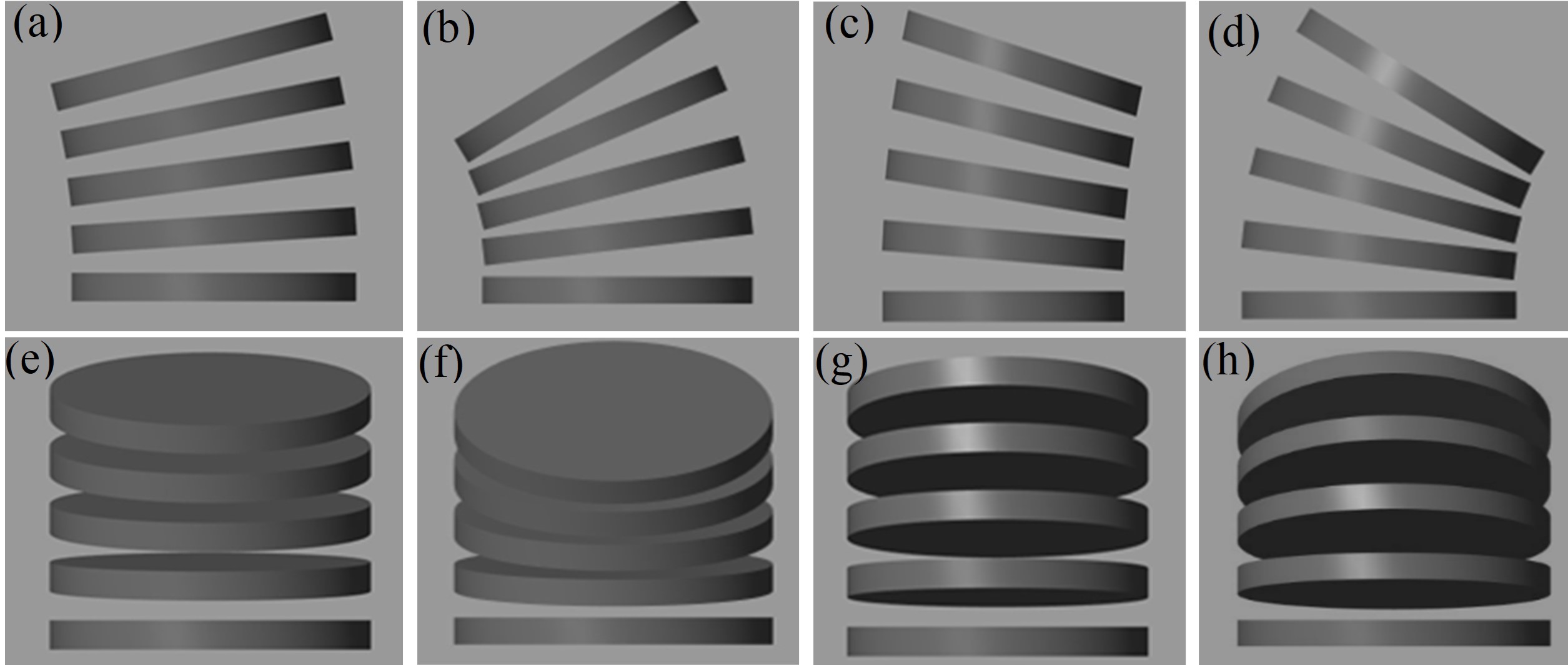}
   \caption{Motion of wrist (a) Radial-1 (b) Radial-2 (c) Ulnar-1 (d) Ulnar-2 (e) Flexion-1 (f) Flexion-2 (g) Extension-1 (h) Extension-2}
    \label{Motion of wrist}
\end{figure}

 Controller loops were created inside Simulink, a MATLAB based software and carried out using a PC with an Intelcore i7 processor and 16 GB RAM. The motions of wrist section in various directions are shown in Figs. \ref{Motion of wrist} (a) - (h). 
The wrist section was assumed to be bending from its initial configuration shown in Figs. \ref{Motion of wrist} to a final bending angle of $30^{0}$ in these directions with respect to the last significant point on disc 5 attached to the hand. The motions of the wrist section were initially controlled using the proposed SMC given in section IV. 
In simulation study, the responses of the system during motion in ulnar deviation direction were only illustrated as all movements were symmetrical, allowing for the prediction of responses in other directions based on the obtained results. 
Damping and stiffness values of the wrist were determined as $0.615$ Nm and $0.105$ Nms respectively.
In the context of soft robots, which may experience significant changes in their mechanical properties due to deformation or environmental interactions, SMC strategy ensures that the robot can still achieve its desired trajectory and performance metrics. 
By effectively managing the non-linearities and uncertainties associated with motions of soft wrist section, SMC facilitated smoother and more efficient motions, thereby improving the overall functionality of the system \cite{ANN5}. Tuning parameters ($P_1$, $P_2$, and $P_3$) were essential for achieving the desired trade-off between system responsiveness and stability. These parameters dictated the sliding surface's characteristics, which in turn aided the system to reject disturbances and track reference signals. If the parameters are not optimally set, the controller can exhibit undesirable behaviors such as chattering, which can lead to wear and tear on mechanical components or even system instability.
Initially $P_1$, $P_2$, and $P_3$ were set to 1. These parameters were tuned using Particle Swarm Optimisation (PSO) to obtain optimal values for reducing output error and chattering. Final values of $P_1$, $P_2$, and $P_3$ were determined as 0.001,1000, and 1000 respectively. 

An ANN block was included in the control scheme for determining desired bending angles. ANN dataset was generated using kinematic model obtained using PCC method. The ANN was trained using a batch size of 100 samples over the course of 100 epochs. The dataset was partitioned such that 75\% was allocated for training purposes, while the remaining 25\% percent was designated for validation of the network's performance. The performances of the networks were compared in table \ref{table:1}. Based on the accuracy of the results, which are evident from table \ref{table:1}, we selected sigmoid function as the activation function.

\begin{table}[hbt!]
    \caption{ANN based training parameters and results}
    \centering
    \begin{tabular}{|c|c|c|c|c|}
    \hline
     Parameters &Sigmoid &ReLU &Leaky ReLU &Tanh \\
         \hline
        Batch size &100 &100 &100 &100\\
        Epochs &100 &100 &100  &100\\
        Gradient &0.0012   &0.097    &0.041  &0.011   \\
        Momentum &0.90   &0.95      &0.92  &0.97   \\
        Weight decay &0.01 &0.01 &0.01 &0.01  \\
        Accuracy &99.7    & 95.5 &97.1  &98.8    \\
        Learning rate &  0.003  &0.001 & 0.001 &0.002 \\
        Training loss &0.05   &0.55 &0.31  &0.22        \\
        Validation loss &0.01 &0.25 &0.10 &0.09  \\
           \hline
    \end{tabular}
    
    \label{table:1}
\end{table}

\begin{figure}
\centerline{\includegraphics[width=0.8\textwidth]{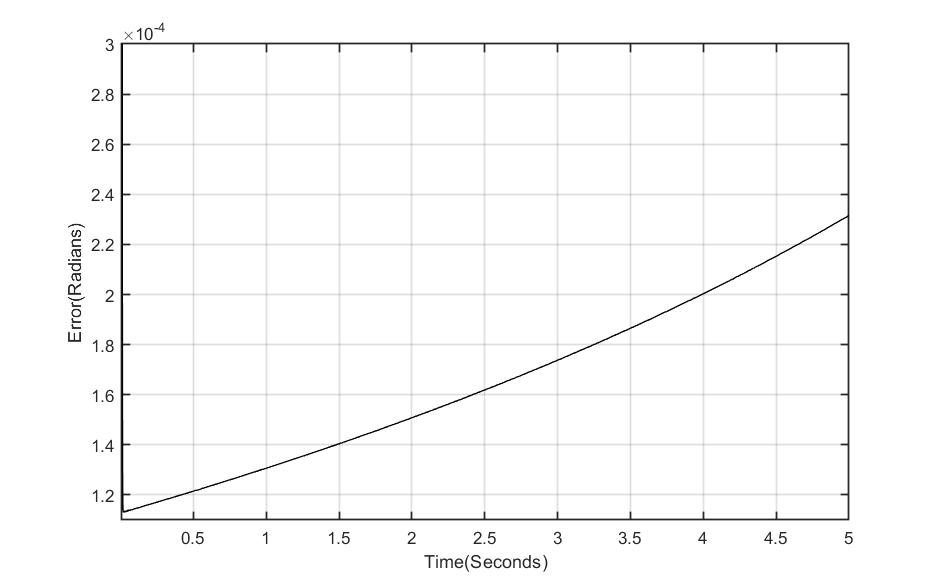}}
\caption{ Error in angles using SMC during ulnar deviation} 
\label{error}
\end{figure}

\begin{figure}[hbt!]
\centerline{\includegraphics[width=1\textwidth]{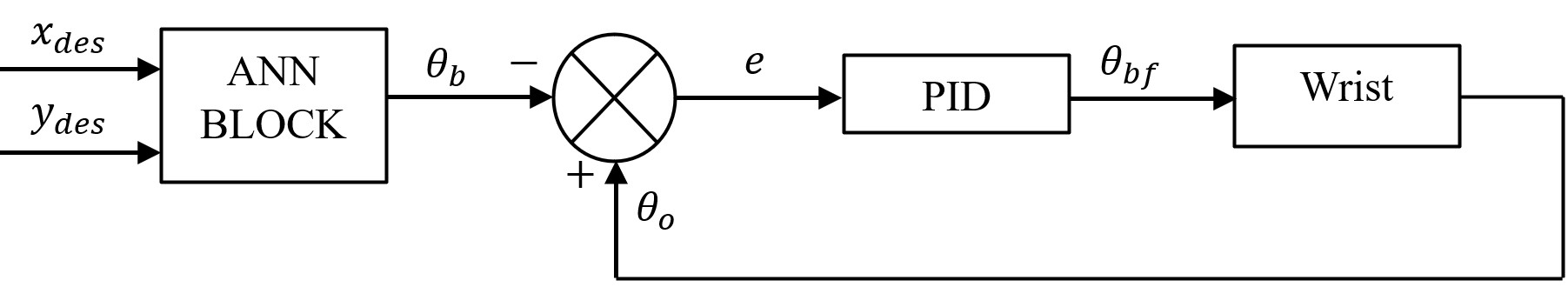}}
\caption{ PID based control strategy} 
\label{pid_block}
\end{figure}

\begin{figure}
\centerline{\includegraphics[width=0.8\textwidth]{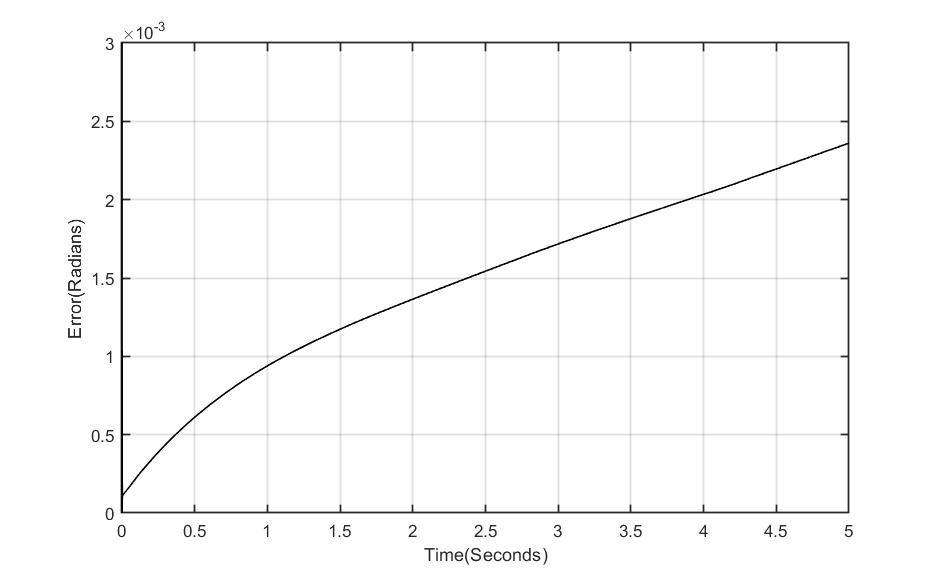}}
\caption{ Error in angles using PID controller during ulnar deviation} 
\label{error_pid}
\end{figure}

The errors in the values of desired bending angles with respect to output bending angles obtained using the SMC strategy during simulation study are shown in fig. \ref{error}. Root Mean Square Error (RMSE) of error values was obtained as $2.7 \times 10^{-4}$ radians. Settling time and steady state error values were determined as $1.2$ s and $2.31 \times 10^{-4}$ radians respectively. The RMSE value, settling time, and steady state error values were obtained within the tolerance limit.
%The comparison of desired and output bending angles of wrist is shown in figure.  
%Figure:error smc
The performance of the SMC strategy was compared with an ANN based PID controller by replacing SMC block with a PID controller (shown in figure \ref{pid_block}). 
The outputs from ANN block were corrected after incorporating error, $e(t)$ and fed to plant model as given in equation \ref{eq:22}
\begin{equation}
\theta_{bf}(t) = K_{P}e(t)+K_{I}\int_{0}^{t} e(t)~dt+K_{D}\frac{de}{dt}
    \label{eq:22}
\end{equation}
%where 
%\begin{equation}
   % e(t)=\theta_o -\theta_b
%\end{equation}
The PID coefficients were determined using crude tuning method and values were obtained as proportional gain, $K_{p} = 1 \times 10^{4} $, integral gain, $K_{i}=5 \times 10^{3}$, and differential gain, $K_{d}=2 \times 10^{3}$. Fig. \ref{error_pid} depicts the errors in the required bending angle values compared to the output bending angles obtained during simulation study using PID based controller. RMSE of error , settling time, and steady state error were computed as $1.5 \times 10^{-3}$ radians, $3$ s, and $2.48 \times 10^{-3}$ radians respectively. These values were higher compared to results obtained using SMC. Hence, SMC demonstrated superior efficiency and performance compared to PID controller.

\subsection{Comparison with other controllers}
The performance of the neuro-controller was also compared with other model based controllers developed for the wrist section. Results obtained using a Model Reference Adaptive Controller (MRAC) \cite{mrac}, a Geometric Variable Strain Controller (GVSC) \cite{conf} and a Neuro Controller (NC) \cite{ref15} developed for the wrist section were compared with the results obtained using the proposed controller. RMSE, settling time and  steady state error were compared to analyse the performances of the controllers as given in table \ref{table:2}. %Results during the motion of wrist in ulnar deviation direction are given in table \ref{table:3}.
\begin{table*}[hbt!]
\caption{Comparison study}
  \centering
    \begin{tabular}{|c|c|c|c|c|}
    \hline 
    Parameters  &MRAC &GVSC &NC \\
    \hline
     RMSE  &$1.2 \times 10^{-3}$  &$2.9 \times 10^{-2}$ &$5.0 \times 10^{-2}$  \\
    \hline 
   
 Settling time (s) &$2.8$  &$3.18$ & $1.02$  \\
    \hline 
    Steady state error  &$1.3 \times 10^{-3}$ &$1.4 \times 10^{-2}$ &$1.9 \times 10^{-2}$   \\
    \hline 
     \end{tabular}\\
  \label{table:2}
\end{table*}
Proposed SMC controller performed better in terms of RMSE and Steady state error. However, settling time of the proposed SMC was higher compared to NC. NC converged faster within 1.02 seconds and performed better in terms of settling time compared to other controllers. However, RMSE and steady state error values were higher compared to other controllers. GVSC approach showcased higher  steady state error, settling time and RMSE values compared to MRAC approach. In terms of RMSE and  steady state error values, MRAC  performed better than GVSC and NC.

\subsection{Experimental validation of SMC}
%introduction

\begin{figure}[hbt!]
\centerline{\includegraphics[width=1\textwidth]{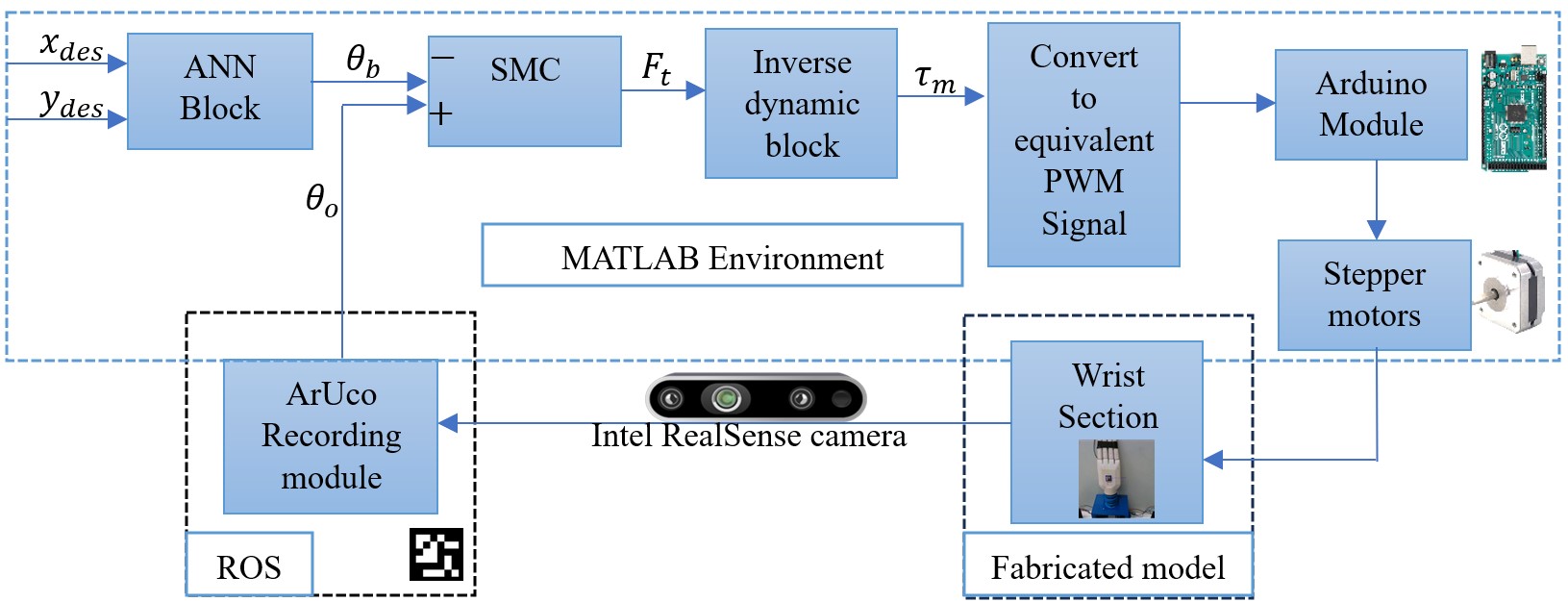}}
\caption{ Flow of data during experimentation } 
\label{data}
\end{figure}

\begin{figure}[hbt!]
\centerline{\includegraphics[width=0.7\textwidth,height = 5 in]{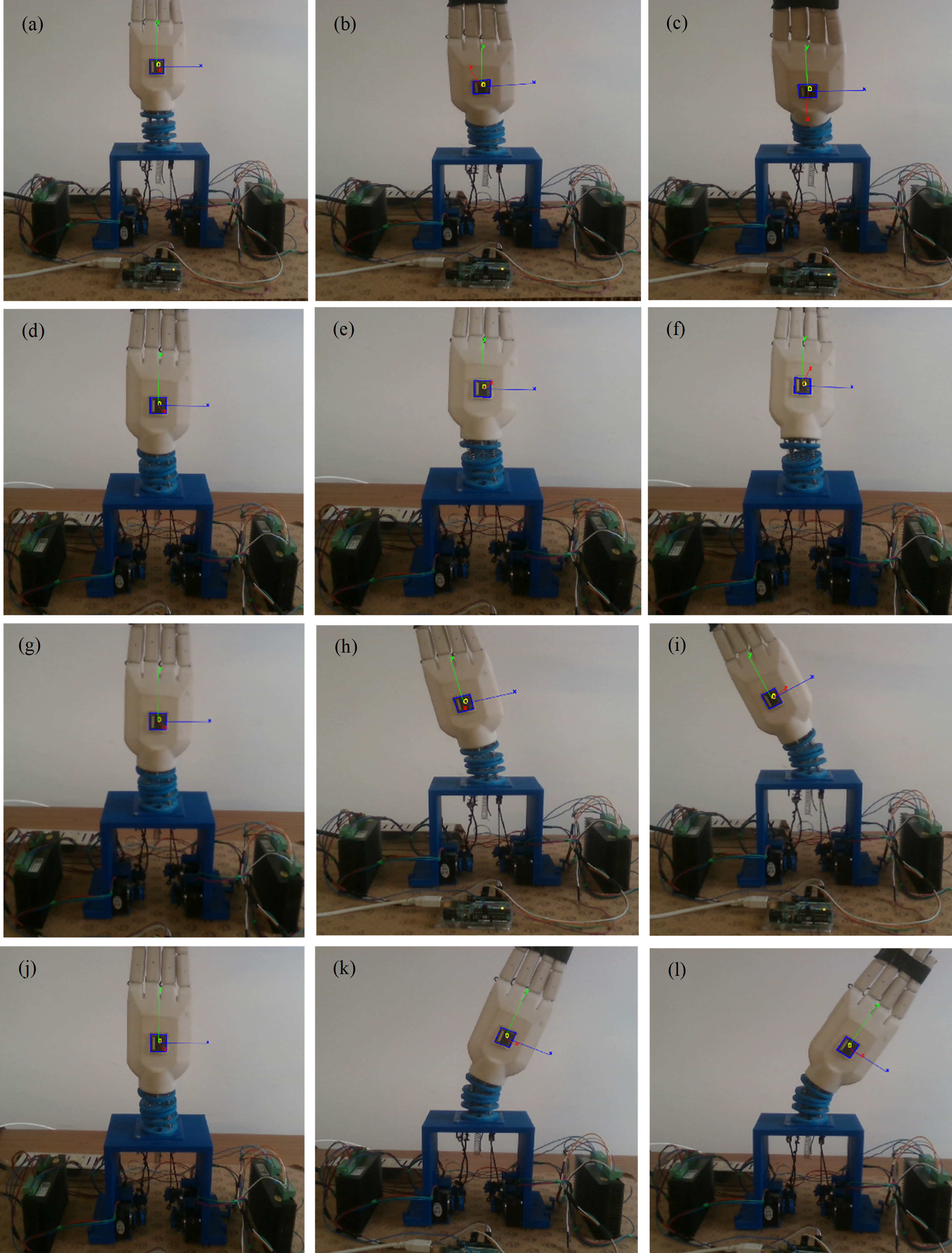}}
\caption{Motion of hand during experimentation (a)-(c)Extension (d)-(f)Flexion (g)-(i)Radial (j)-(l)Ulnar} 
\label{error_exp}
\end{figure}

\begin{figure}[hbt!]
\centerline{\includegraphics[width=01\textwidth, height = 4 in]{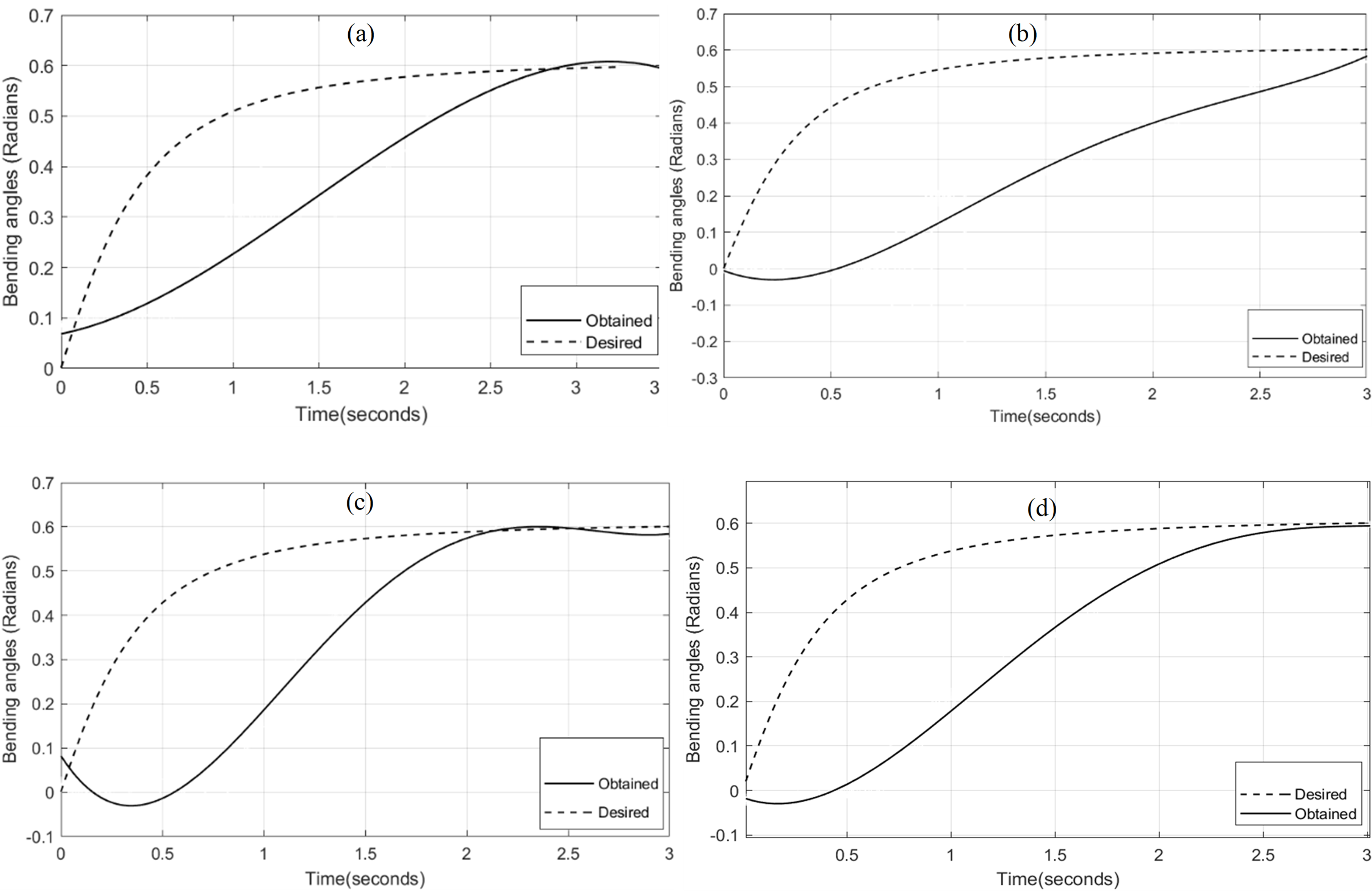}}
\caption{Comparison of desired and obtained bending angles during experimentation(a)Extension (b)Flexion (c)Radial (d)Ulnar} 
\label{exp1}
\end{figure}

\begin{figure}[hbt!]
\centerline{\includegraphics[width=1\textwidth, height = 4 in]{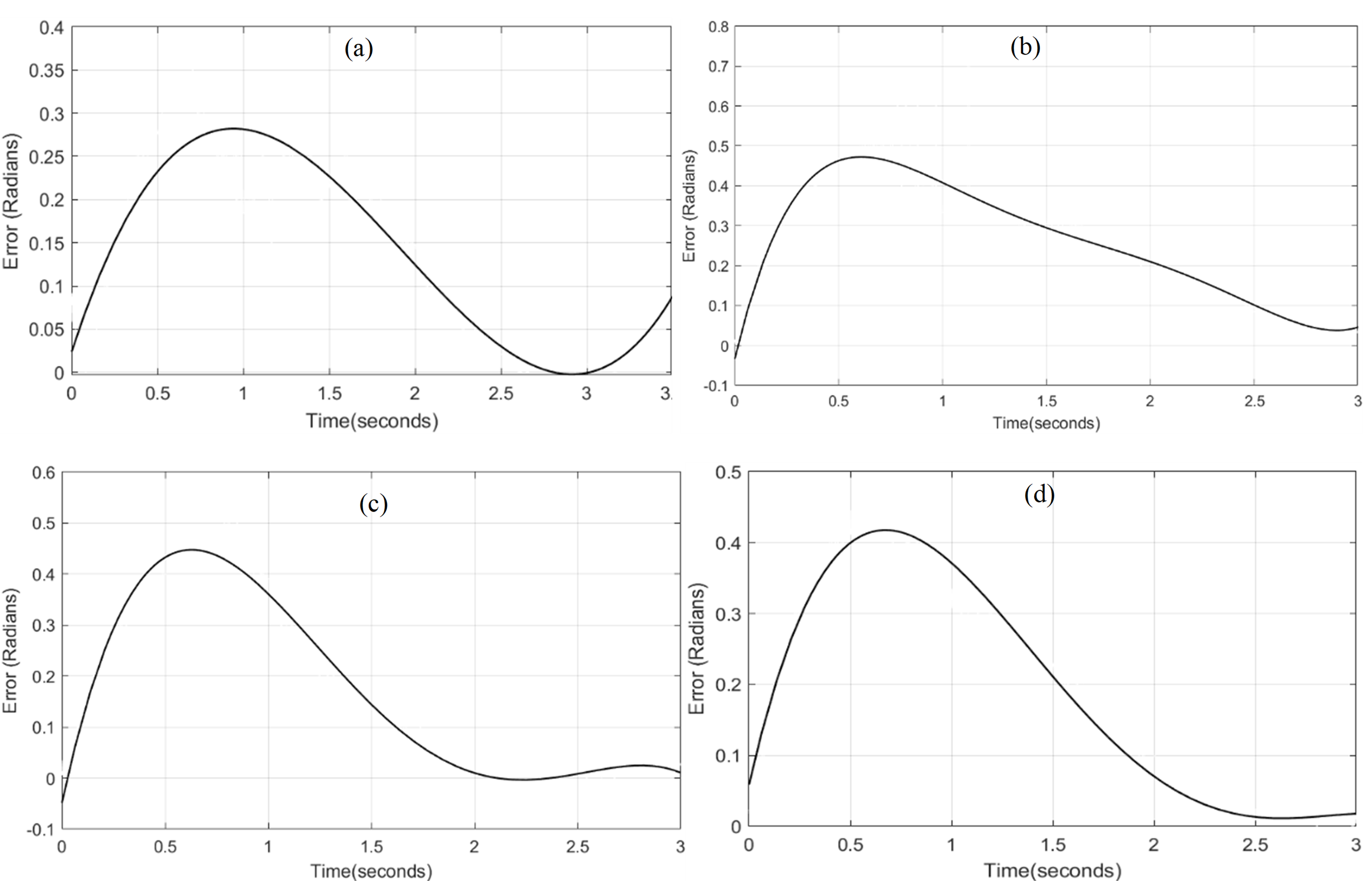}}
\caption{Error in hand motions during experimentation(a)Extension (b)Flexion (c)Radial (d)Ulnar} 
\label{exp2}
\end{figure}

The validation of the proposed controller scheme and its motion capabilities were conducted using a fabricated wrist section integrated to the PRISMA hand II (shown in Fig. \ref{conceptual model}(b)). Four peripheral tendons were utilized to enable rotational movements in four separate directions. Tendons 1 and 2 were engaged to facilitate radial deviation of the wrist, while tendons 4 and 5 managed movements in the ulnar direction. Furthermore, tendons 1 and 4 were employed for extension motions, whereas flexion was governed by tendons 2 and 5. The lowest disc (disc 1) was secured to a stable platform, and the highest disc (disc 5) was connected to the hand. Tendons 4 and 5 were pulled by actuating motors 3 and 4 in anticlockwise direction to rotate the wrist in ulnar direction. An ArUco marker placed on the wrist was used for tracking motions of the disc 5 during experimentation. Desired values of motion were given as the input to the ANN block. Desired bending angles obtained from ANN block were given as the inputs to SMC. The desired actuation forces generated from SMC block were converted to motor torque values and later converted to PWM signals. The PWM signals were fed to an Arduino Mega 2560. The Arduino was connected to 2 stepper motors using a microstep driver (DM542). 
An Arduino module was integrated with Simulink to facilitate the transmission of the calculated input signals and a ROS module was employed to relay the readings from the ArUco markers into the Simulink environment. The flow of data during experimentation is shown in Fig. \ref{data}. 
The motions of wrist carrying the hand during experimentation are shown in Figs. \ref{error_exp} (a) - (l). Comparison of desired and obtained bending angles during experimentation is shown in Fig \ref{exp1} (a) - (d).  
During experimentation, average values of RMSE of error values of bending angles, settling time, and steady state error were obtained as 0.157 radians, 3 s and 0.032 radians respectively. The errors during the hand motion are shown in Fig. \ref{exp2}. 
The data clearly indicates that the error values observed during experimentation were significantly greater than those recorded during the simulation study. The primary factor contributing to the
increased error margin observed during the experimental
phase was the reduced stiffness of the springs utilized in the wrist segment. Nevertheless, it is noteworthy that the wrist successfully supported the payload while ensuring the stability of the hand throughout the motions.

\begin{figure}[hbt!]
\centerline{\includegraphics[width=0.7\textwidth,height = 3 in]{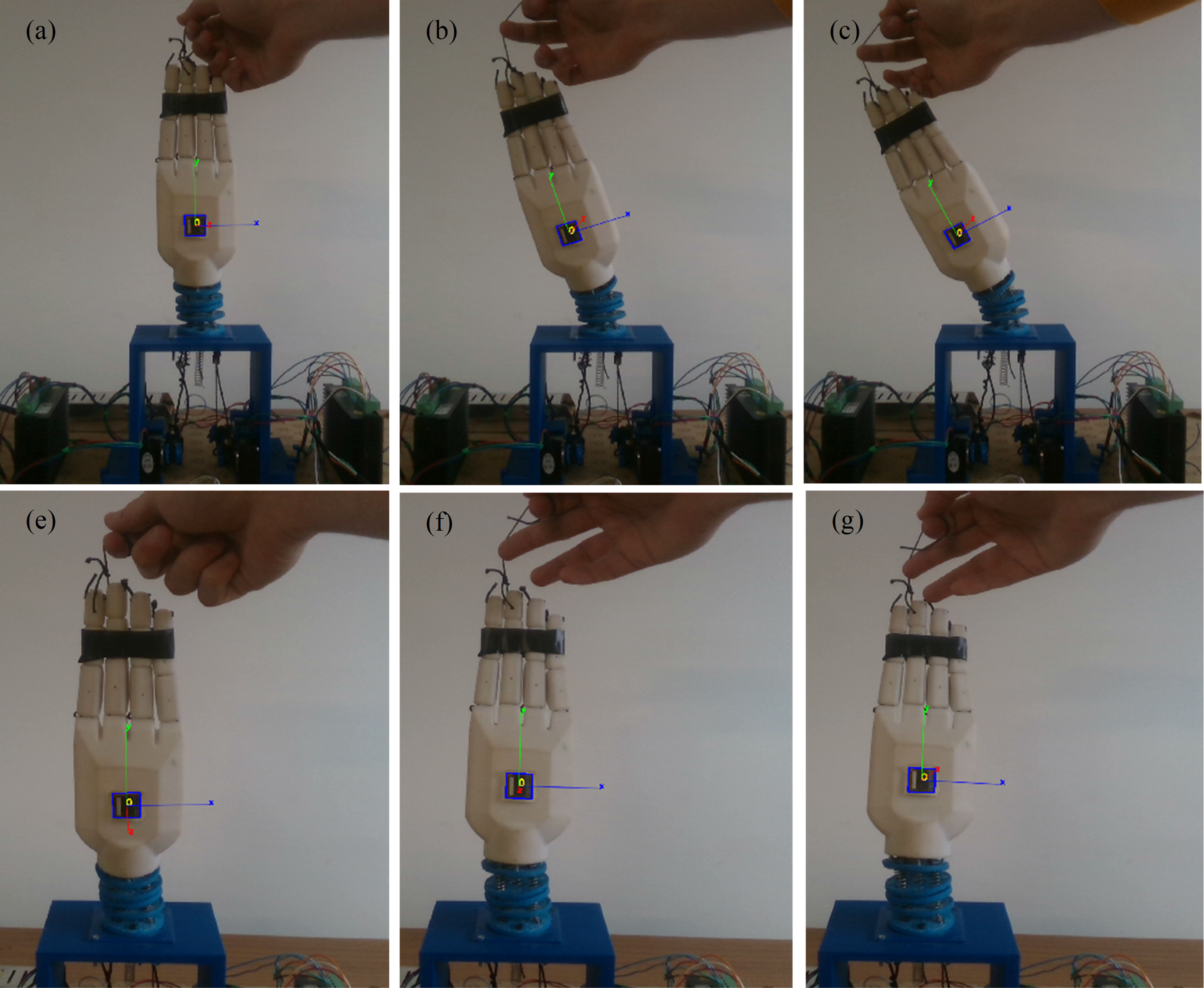}}
\caption{ Errors in the hand motion during experimentation in presence of external disturbances (a) - (c)Radial (e) - (g)Flexion } 
\label{error_exp1}
\end{figure}

\begin{figure}[hbt!]
\centerline{\includegraphics[width=1\textwidth,height = 4.2 in]{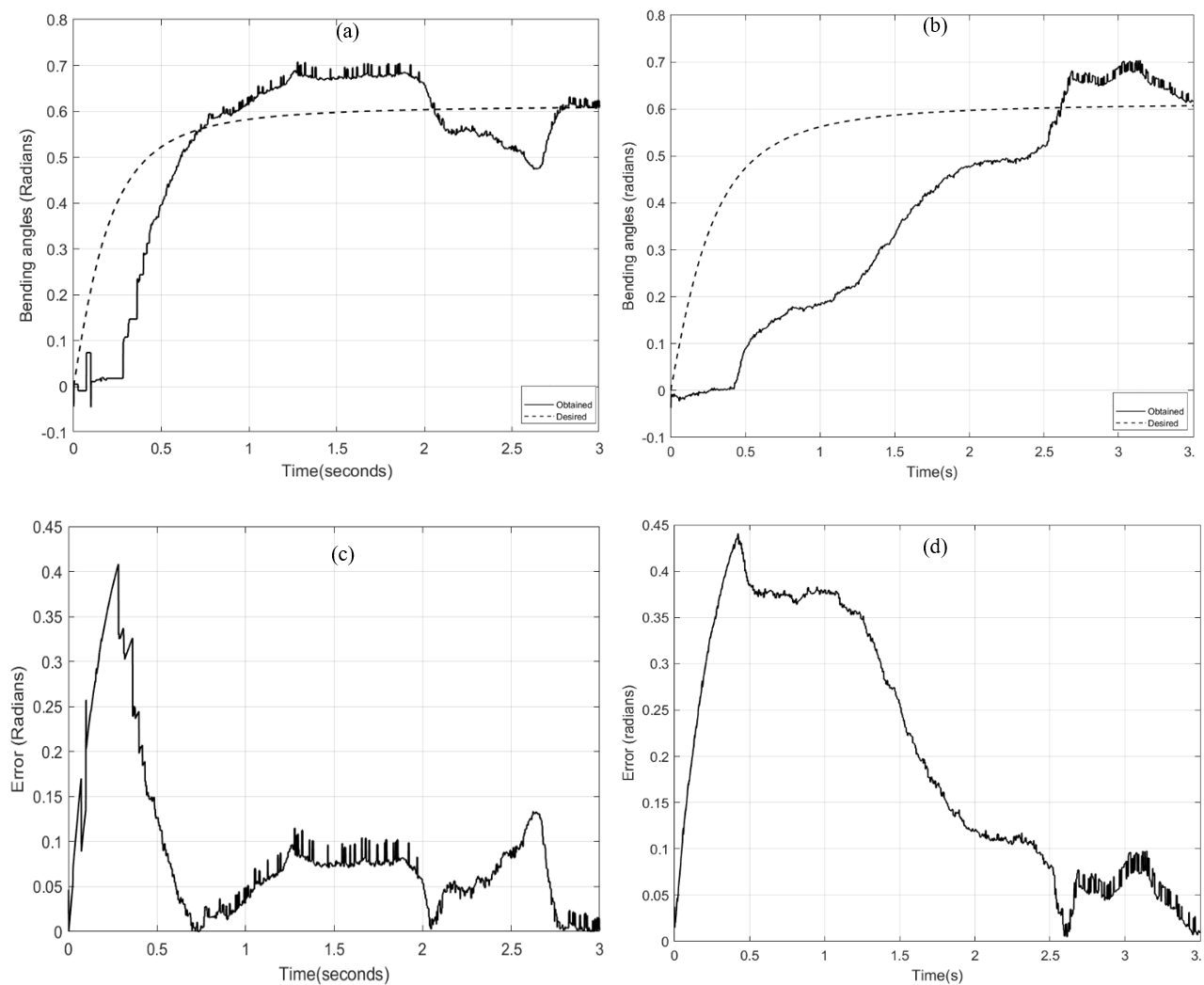}}
\caption{ Comparison of desired and obtained bending angles (a)Extension (b)Flexion (c)Radial (d)Ulnar} 
\label{bending angles}
\end{figure}

To assess the controller's adaptability in response to external unknown forces, an external force was applied using a human hand while the prosthetic hand executed radial deviation and flexion movements, as shown in Figs. \ref{error_exp1} (a) - (g). The comparison of desired and obtained bending angles during the application of external disturbances are presented in Figs. \ref{exp2} (e) - (f).The motion errors during the application of external disturbances are presented in Figs. \ref{bending angles} (e) - (f). 

The motion showcased an average RMSE value of 0.207 m in radial direction and RMSE value of 0.291 m in flexion direction. Average value of settling time for radial and flexion directions was obtained as 3 seconds. Steady state error values were obtained as 0.029 s for radial direction and 0.031 s for flexion direction. The values of settling time, steady state error and RMSE were higher as compared to experimental results without the presence of force. However, these values are in tolerance range and the convergence of the results indicates the controller's capability to manipulate the motions of wrist section even in the presence of external disturbances.

\section{CONCLUSION AND FUTURE WORKS}

Development of a sophisticated control framework is essential for optimizing the performance of soft continuum robots. 
%These robots, characterized by their flexible and deformable structures, require a control system that can accommodate their unique kinematics and dynamics. By implementing advanced algorithms and feedback mechanisms, we can enhance the robot's responsiveness and stability during operation.
This work presented a PCC model-based SMC, that can attain dynamic control of a soft continuum wrist integrated to a prosthetic hand. The integration of ANN with SMC contributed several significant advantages that enhanced system performance and robustness. By employing the SMC approach, the system maintained desired motion trajectories even in the presence of variations in the robot's physical properties and environmental conditions.The alignment of the results during the application of external disturbances demonstrated the controller's ability to regulate the movements of the wrist section, even when faced with external disturbances. ANN contributed the controller performance by reducing computational time. The performance of the SMC was compared with an ANN based PID controller and other model based controllers. The SMC exhibited a lower RMSE of errors compared to a PID controller using simulation studies. Furthermore, values of settling time and steady state errors exhibited by SMC were lower compared to PID controller. Hence, using the simulation study, it was concluded that SMC performed better compared to the PID controller. The proposed SMC controller also demonstrated superior performance with respect to RMSE and steady-state error compared to model based controllers. During experimentation, RMSE values of errors were observed higher compared to simulation values due to spring stiffness variation. 
Future works will focus on re-fabricating the current design of wrist to create a more robust structure. Controller strategies will be updated by integrating real-time sensor feedback to increase the accuracy of motion. 

\section*{Declarations:}

\section*{Funding}
This work was supported by the Italian Ministry of Research under the complementary actions to the NRRP ”Fit4MedRob - Fit for Medical Robotics” Grant (PNC0000007).

\section*{Conflict of interest/Competing interests}
All authors declare that they have no conflicts of interest.

\section*{Ethics approval and consent to participate
}
Not applicable

\section*{Consent for publication}
Not applicable

\section*{Data availability statement}
The data associated with this work will be shared by contacting the authors.

\section*{Materials availability}
Not applicable
\section*{Code availability}
The code will be made available by contacting the corresponding author.
\section*{Author contribution}
Shifa Sulaiman derived the mathematical models, performed analytic calculations, carried out simulations, experimental validations, and wrote the paper. Francesco Schetter contributed data, analysis tools and assisted in carrying out experimental validations. Mehul Menon performed a part of the simulation study. Fanny Ficuciello investigated and supervised the work.

%%%%%%%%%%%%%%%%%%%%%%%%%%%%%%%%%%%%%%%%%%%%%%%%%%%%%%%%%%%%%%%%%%%%%%%%%%%%%%%%

%References are important to the reader; therefore, each citation must be complete and correct. If at all possible, references should be commonly available publications.\cite{c1}


\begin{thebibliography}{99}

\bibitem{fanny1}F. Ficuciello, "Synergy-Based Control of Underactuated Anthropomorphic Hands," in IEEE Transactions on Industrial Informatics, vol. 15, no. 2, pp. 1144-1152, Feb. 2019, doi: 10.1109/TII.2018.2841043.

\bibitem{fanny2}Villani, L., Ficuciello, F., Lippiello, V., Palli, G., Ruggiero, F., Siciliano, B. (2012). Grasping and Control of Multi-Fingered Hands. In:Siciliano, B. (eds) Advanced Bimanual Manipulation. Springer Tracts in Advanced Robotics, vol 80. Springer, Berlin, Heidelberg. https://doi.org/10.1007/978-3-642-29041-15.

\bibitem{de}de Souza JO, Bloedow MD, Rubo FC, de Figueiredo RM, Pessin G, Rigo SJ. Investigation of different approaches to real-time control of prosthetic hands with electromyography signals. IEEE Sensors Journal. 2021 Jul 26;21(18):20674-84.

\bibitem{control1}
Sulaiman, S., Schetter, F., Shahabi, E., \& Ficuciello, F. (2025). A learning based impedance control strategy implemented on a soft prosthetic wrist in joint-space. Frontiers in Robotics and AI, 12, 1665267.

\bibitem{control2}
Sulaiman, S., Gohari, M., Schetter, F., \& Ficuciello, F. (2026, January). A Learning-Based Model Reference Adaptive Controller Implemented. In Social Robotics+ AI: 17th International Conference, ICSR+ AI 2025, Naples, Italy, September 10–12, 2025, Proceedings, Part I (p. 243). Springer Nature.

\bibitem{palli}Palli G, Melchiorri C, Vassura G, et al. The DEXMART hand: Mechatronic design and experimental evaluation of synergy-based control for human-like grasping. The International Journal of Robotics Research. 2014;33(5):799-824. doi:10.1177/0278364913519897.

\bibitem{palli1}Fanny Ficuciello, Gianluca Palli, Claudio Melchiorri, Bruno Siciliano, Postural synergies of the UB Hand IV for human-like grasping, Robotics and Autonomous Systems, Volume 62, Issue 4, 2014,Pages 515-527, ISSN 0921-8890, https://doi.org/10.1016/j.robot.2013.12.008.

\bibitem{review}Armanini C, Boyer F, Mathew AT, Duriez C, Renda F. Soft robots modeling: A structured overview. IEEE Transactions on Robotics. 2023 Jan 6;39(3):1728-48.

%\bibitem{CNN}Chen X, Li Y, Hu R, Zhang X, Chen X. Hand gesture recognition based on surface electromyography using convolutional neural network with transfer learning method. IEEE Journal of Biomedical and Health Informatics. 2020 Jul 15;25(4):1292-304.



%\bibitem{orekhov2017modeling} A. L. Orekhov, V. A.  Aloi, D. C. Rucker, "Modeling parallel continuum robots with general intermediate constraints", In 2017 IEEE International Conference on Robotics and Automation (ICRA) 2017 May 29 (pp. 6142-6149). IEEE.

%\bibitem{wooten2018vine}M. B. Wooten, I.D. Walker, "Vine-inspired continuum tendril robots and circumnutations", Robotics. 2018 Sep 18;7(3):58.
\bibitem{ANN}
Sulaiman, S., Schetter, F., Shahabi, E., \& Ficuciello, F. (2025). A learning based impedance control strategy implemented on a soft prosthetic wrist in joint-space. Frontiers in Robotics and AI, 12, 1665267.

\bibitem{wockenfuss2022design} W. R. Wockenfuß, V. Brandt, L. Weisheit, W. G. Drossel, "Design, modeling and validation of a tendon-driven soft continuum robot for planar motion based on variable stiffness structures", IEEE Robotics and Automation Letters. 2022 Feb 7;7(2):3985-91.

\bibitem{tutcu2021quasi} C. Tutcu,  B. A. Baydere , S. K. Talas, E. Samur, "Quasi-static modeling of a novel growing soft-continuum robot", The International Journal of Robotics Research. 2021 Jan;40(1):86-98.

\bibitem{escande}C. Escande, T. Chettibi, R. Merzouki, V. Coelen, and P. M. Pathak, “Kinematic calibration of a multisection bionic manipulator,” IEEE/ASME
Trans. Mechatron., vol. 20, no. 2, pp. 663–674, Apr. 2015.

\bibitem{huang}Huang X, Zou J, Gu G. Kinematic modeling and control of variable curvature soft continuum robots. IEEE/ASME Transactions on Mechatronics. 2021 Jan 28;26(6):3175-85.

\bibitem{xavier}M. Xavier, A. Fleming, and Y. Yong, “Finite element modeling of soft
fluidic actuators: Overview and recent developments,” Adv. Intell. Syst.,
vol. 3, no. 2, 2021, Art. no. 2000187

\bibitem{renda}M. Giorelli, F. Renda, M. Calisti, A. Arienti, G. Ferri, and C. Laschi,
“Neural network and jacobian method for solving the inverse statics of a
cable-driven soft arm with nonconstant curvature,” IEEE Trans. Robot.,
vol. 31, no. 4, pp. 823–834, Aug. 2015


\bibitem{ANN1}
Della Santina, C., Bicchi, A., \& Rus, D. (2020). On an improved state parametrization for soft robots with piecewise constant curvature and its use in model based control. IEEE Robotics and Automation Letters, 5(2), 1001-1008.

\bibitem{r1}H. D. Yang and A. T. Asbeck, “Design and characterization of a modular
hybrid continuum robotic manipulator,” IEEE/ASME Transactions on
Mechatronics, vol. 25, no. 6, pp. 2812–2823, 2020.

\bibitem{r2}S. Neppalli and B. A. Jones, “Design, construction, and analysis of
a continuum robot,” in 2007 IEEE/RSJ International Conference on
Intelligent Robots and Systems. IEEE, 2007, pp. 1503–1507.

\bibitem{rao}Rao P, Pogue C, Peyron Q, Diller E, Burgner-Kahrs J. Modeling and analysis of tendon-driven continuum robots for rod-based locking. IEEE Robotics and Automation Letters. 2023 Apr 5;8(6):3126-33.

\bibitem{cheng}Cheng H, Liu H, Wang X, Liang B. Approximate piecewise constant curvature equivalent model and their application to continuum robot configuration estimation. In2020 IEEE International Conference on Systems, Man, and Cybernetics (SMC) 2020 Oct 11 (pp. 1929-1936). IEEE.

%\bibitem{della}Della Santina C, Katzschmann RK, Biechi A, Rus D. Dynamic control of soft robots interacting with the environment. In2018 IEEE International Conference on Soft Robotics (RoboSoft) 2018 Apr 24 (pp. 46-53). IEEE.



%\bibitem{doroudchi2021configuration}A. Doroudchi, and S. Berman, "Configuration tracking for soft continuum robotic arms using inverse dynamic control of a cosserat rod model". In 2021 IEEE 4th International Conference on Soft Robotics (RoboSoft), 2021, (pp. 207-214). IEEE.

%\bibitem{levant}Levant A. Sliding order and sliding accuracy in sliding mode control. International journal of control. 1993 Dec 1;58(6):1247-63.

%\bibitem{Kumar}Kumar A, Anwar MN, Kumar S. Sliding mode controller design for frequency regulation in an interconnected power system. Protection and Control of Modern Power Systems. 2021 Jan;6(1):1-2.
\bibitem{ANN2}
George Thuruthel, T., Ansari, Y., Falotico, E., \& Laschi, C. (2018). Control strategies for soft robotic manipulators: A survey. Soft robotics, 5(2), 149-163.

\bibitem{ANN3}
Gohari, M., Sulaiman, S., Schetter, F., \& Ficuciello, F. (2025, February). A Sliding Mode Controller Design Based on Timoshenko Beam Theory Developed for a Prosthetic Hand Wrist. In 2025 11th International Conference on Automation, Robotics, and Applications (ICARA) (pp. 338-342). IEEE.

\bibitem{cao}Cao G, Liu Y, Jiang Y, Zhang F, Bian G, Owens DH. Observer-based continuous adaptive sliding mode control for soft actuators. Nonlinear Dynamics. 2021 Jul;105(1):371-86.
\bibitem{Kazemipour}Kazemipour A, Fischer O, Toshimitsu Y, Wong KW, Katzschmann RK. Adaptive dynamic sliding mode control of soft continuum manipulators. In2022 International Conference on Robotics and Automation (ICRA) 2022 May 23 (pp. 3259-3265). IEEE.
\bibitem{Skorina}Skorina EH, Luo M, Ozel S, Chen F, Tao W, Onal CD. Feedforward augmented sliding mode motion control of antagonistic soft pneumatic actuators. In2015 IEEE International Conference on Robotics and Automation (ICRA) 2015 May 26 (pp. 2544-2549). IEEE.

\bibitem{khan}Khan AH, Li S. Sliding mode control with PID sliding surface for active vibration damping of pneumatically actuated soft robots. IEEE access. 2020 May 7;8:88793-800.

\bibitem{Mousa}Mousa A, Khoo S, Norton M. Robust control of tendon driven continuum robots. In2018 15th International workshop on variable structure systems (VSS) 2018 Jul 9 (pp. 49-54). IEEE.

\bibitem{Al}Alqumsan AA, Khoo S, Arogbonlo A, Nahavand S. Adaptive neural network based sliding mode control of continuum robots with mismatched uncertainties. In2021 IEEE International Conference on Systems, Man, and Cybernetics (SMC) 2021 Oct 17 (pp. 2602-2607). IEEE.

%\bibitem{Wu}Wu Q, Wang Z, Chen Y. sEMG-based adaptive cooperative multi-mode control of a soft elbow exoskeleton using neural network compensation. IEEE Transactions on Neural Systems and Rehabilitation Engineering. 2023 Aug 17.




%\bibitem{Guan}Guan M, Qu C, Lv J, Yang L, Li X. A novel RBF neural network–based sliding mode controller for a master–slave motor coordinated drive system. The International Journal of Advanced Manufacturing Technology. 2024 Jun 26:1-5.

\bibitem{conf}
Sulaiman, Shifa, Mehul Menon, Francesco Schetter, and Fanny Ficuciello. "Design, Modelling, and Experimental Validation of a Soft Continuum Wrist Section Developed for a Prosthetic Hand." In 2024 IEEE/RSJ International Conference on Intelligent Robots and Systems (IROS), pp. 11347-11354. IEEE, 2024.

\bibitem{ref2}
Liu H, Ferrentino P, Pirozzi S, Siciliano B, Ficuciello F. The PRISMA Hand II: a sensorized robust hand for adaptive grasp and in-hand manipulation. In ISRR. 2019 pp. 971-986. Cham: Springer International Publishing.

\bibitem{della}Della Santina C, Katzschmann RK, Biechi A, Rus D. Dynamic control of soft robots interacting with the environment. In2018 IEEE International Conference on Soft Robotics (RoboSoft) 2018 Apr 24 (pp. 46-53). IEEE.

\bibitem{ANN4}
Yang, J., \& Long, Q. (2024). A modification of adaptive moment estimation (adam) for machine learning. Journal of Industrial and Management Optimization, 20(7), 2516-2540.

\bibitem{ANN5}
Kapoor, N., \& Ohri, J. (2017). Sliding mode control (SMC) of robot manipulator via intelligent controllers. Journal of The Institution of Engineers (India): Series B, 98(1), 83-98.

\bibitem{mrac}S. Sulaiman, P. De Risi, F. Schetter and F. Ficuciello, "A Stable Model Reference Adaptive Controller Developed for a Prosthetic Hand Wrist," in IEEE Transactions on Automation Science and Engineering, vol. 22, pp. 24413-24434, 2025, doi: 10.1109/TASE.2025.3634193.

\bibitem{ref15}S. Sulaiman, M. Gohari, F. Schetter, and F. Ficuciello. "An Adaptive Neuro-Controller Developed for a
Prosthetic Hand Wrist." In  19th International Conference on Intelligent Autonomous Systems 2025.





\end{thebibliography}
\end{document}